 \documentclass[tablecaption=bottom,wcp]{jmlr} 

\usepackage{booktabs}
\usepackage[load-configurations=version-1]{siunitx} 

\jmlrproceedings{AABI 2020}{3rd Symposium on Advances in Approximate Bayesian Inference, 2020}

\title{Factorized Gaussian Process Variational Autoencoders}
  \author{\Name{Metod Jazbec}\nametag{\thanks{Equal contribution.}} \Email{jazbec.metod@gmail.com}\\
  \addr ETH Z\"urich
  \AND
  \Name{Michael Pearce}\footnotemark[1] \Email{scrambledpie@gmail.com}\\
  \addr Warwick University
  \AND
  \Name{Vincent Fortuin}\footnotemark[1] \Email{fortuin@inf.ethz.ch}\\
  \addr ETH Z\"urich
 }

\newcommand{\N}{\mathcal{N}}

\newcommand{\R}{\mathbb{R}}

\newcommand{\E}{\mathbb{E}}

\newcommand{\bigO}{\mathcal{O}}

\usepackage{xspace}

\newcommand{\ourmodel}{FGP-VAE\xspace}

\renewcommand{\epsilon}{\ensuremath\varepsilon}

\newcommand{\zb}{\textbf{z}} 
\newcommand{\xb}{\textbf{x}}
\newcommand{\yb}{\textbf{y}}

\newcommand{\Zb}{\textbf{Z}} 
\newcommand{\Xb}{\textbf{X}}
\newcommand{\Yb}{\textbf{Y}}
\newcommand{\Kb}{\textbf{K}} 
\newcommand{\Ib}{\textbf{I}}

\begin{document}

\maketitle

\begin{abstract}

Variational autoencoders often assume isotropic Gaussian priors and mean-field posteriors, hence 
do not exploit structure in scenarios where we may expect similarity or consistency across latent variables.
Gaussian process variational autoencoders alleviate this problem through the use of a latent Gaussian process, but lead to a cubic inference time complexity.
We propose a more scalable extension of these models by leveraging the independence of the auxiliary features, which is present in many datasets.
Our model factorizes the latent kernel across these features in different dimensions, leading to a significant speed-up (in theory and practice), while empirically performing comparably to existing non-scalable approaches.
Moreover, our approach allows for additional modeling of global latent information and for more general extrapolation to unseen input combinations.

\end{abstract}

\section{Introduction}
\label{sec:intro}


Variational autoencoders (VAEs) have achieved great success in many representation
learning tasks \citep{Kingma2014Auto-encodingBayes, rezende2014stochastic}. However,
their isotropic Gaussian prior and variational posterior hinge on the strong assumption
that all data points are independent. This can often lead to problems in real-world
use cases, where the data exhibit significant correlations
\citep{fraccaro2017disentangled, krishnan2016structured}.

Many alternative priors and posteriors have been proposed
for VAEs \citep{Tomczak2018VAEVampprior, Fortuin2019Som-Vae:Series, kopf2019mixture}.
Especially when each input comes with meta-data, or \emph{auxiliary features},
such extra information can be used to construct Gaussian processes (GPs) and use them as priors in the latent space
\citep{Casale2018GaussianAutoencoders, Fortuin2019GP-VAE:Imputation, Pearce2019ThePixels}.
By choosing appropriate kernels, these resulting GP-VAE models allow to capture the structured correlations across latent variables
of different data points \citep{Rasmussen2006GaussianLearning}. 
However, they are also haunted by the computational cost of exact GP
inference, which scales as $\bigO(N^3)$ for $N$ elements in the dataset.

We consider the setting in which the inputs are images, and each image is 
associated with several features, some of which are unique to
the image and some of which are shared with other images.
For example in a set of MNIST digits rotated by multiple angles
\citep{Casale2018GaussianAutoencoders}, the digit ID 
is shared by other images of the same digit while the angle is unique.
Similarly, in a set of faces viewed from multiple perspectives, the person is common to
multiple images while each image has a unique position
\citep{Casale2018GaussianAutoencoders}. Further applications include a set of scenes
viewed from multiple positions \citep{AliEslami2018NeuralRendering}, high dimensional spatio-temporal datasets \citep{ashman2020sparse},
or speech segments where one speaker's voice is shared by
multiple segments \citep{li2018disentangled}.
%
%

For such settings, we propose a novel \emph{factorized} GP-VAE model, \ourmodel, with two desirable properties. 
Firstly, by carefully exploiting factorization, inference is significantly reduced from $\bigO(N^3)$.
Secondly, the representations are encouraged to be disentangled, which is a highly desirable property for VAE models 
\citep{locatello2019fairness, van2019disentangled, trauble2020independence}.

We describe the problem setting in Section \ref{sec:problem_setting} and the proposed model in Section~\ref{sec:method}. We present experimental results in Section~\ref{sec:experiments}, and conclude in Section~\ref{sec:conclusion}.

\section{Problem Setting} \label{sec:problem_setting}

Consider high-dimensional data of $N$ elements
$\Yb = [\yb_1, \dots, \yb_N]^\top$ where $ \yb_i\in \mathbb{R}^{K}$ and each data
point has corresponding low-dimensional auxiliary data 
$\Xb = [\xb_1, \dots, \xb_N]^\top \in \mathcal{X}^{N}, \mathcal{X}\subseteq \R^D$. 
For ease of exposition, we will focus on the example of the rotated MNIST dataset \citep{Casale2018GaussianAutoencoders}.
%
%
It consists of $P$ digits, each observed at $Q$ different angles, amounting to
a total of $N = P \cdot Q$ images.\footnote{The assumption that all digit instances are observed in the same number of angles Q is made to simplify notation; the presented approach
does not rely on this assumption.} Each $\xb_i = (d_i, \: w_i) \in \mathcal{X}$
is composed of a categorical digit instance $d_i$ (integer index or one-hot encoding) and a continuous angle $w_i$. 
%
%
We wish to train a model that can (1) given new $\xb_* \in \mathcal{X}$ generate $\yb_* \in \R^K$,  and (2) infer an interpretable and disentangled latent representation.
 \section{Method}
 \label{sec:method}
 
\textbf{Generative model}: we follow a latent GP approach, first proposed in \cite{Casale2018GaussianAutoencoders} and later extended in \cite{Pearce2019ThePixels}. As in a standard VAE, each image $\yb_i$ is associated with a latent variable $\zb_i \in \R^L$. Making use of the auxiliary data, we further expect that two images $\yb_i$ and $\yb_j$ with similar $\xb_i$ and $\xb_j$ should also have similar $\zb_i$ and $\zb_j$. To this end, a Gaussian process regression is used to model a joint distribution over all latent variables $\Zb = [\zb_1,\dots,\zb_N]^T\in\R^{N\times L}$. Given $L$ latent dimensions, we assume $L$ independent latent functions $f^l \sim GP(0, \: k_{\theta}^{l}), \: l=1, \dots, L$
with kernels $k_\theta^l$ and therefore the latent variable for $\yb_i$ may be written as $\zb_i = [f^1(\xb_i),\dots,f^L(\xb_i)]^T$.
Likewise, all latent variables of the $l^{th}$ channel $\zb^{l}_{1:N} =[f^l(\xb_1),\dots,f^l(\xb_N)] \in\R^N$
are assumed to come from a single (unknown) function, specifically $\zb^{l}_{1:N}$ has a correlated Gaussian prior with covariance $\Kb^l_{NN}=k_\theta^l(\Xb, \Xb) \in \R^{N \times N}$. The generative model,  $p_\psi(\Yb,\Zb|\Xb)=p_\psi(\Yb|\Zb)p_\theta(\Zb|\Xb)$, is thus
\begin{align*}
     p_\theta(\Zb|\Xb) &= \prod_{l=1}^L\N(\zb_{1:N}^{l}| 0, \textbf{K}_{NN}^l), \\
    p_\psi(\Yb|\Zb) &= \prod_{i=1}^Np_\psi(\yb_i|\zb_i) = \prod_{i=1}^N\N(\yb_i| \mu_\psi(\zb_i), \sigma_y^2 \: \Ib_K),
\end{align*}
where $\mu_{\psi} : \R^L \to \R^K$ is a (generative) network with parameters $\psi$. Note that if $\textbf{K}_{NN}=\textbf{I}$ is the identity matrix, the model recovers a standard VAE; using the $\Xb$ values enables the use of more sophisticated prior.

In prior work \citep{Casale2018GaussianAutoencoders, jazbec2020scalable}, a single GP prior is used in all $L$ latent channels. Specifically, for rotated MNIST, a product kernel between a periodic and a linear kernel is considered
\begin{align*}
    k_{\theta}(\xb_i, \xb_j) = \Sigma_{d_i, d_j} \cdot \exp\bigg(-\frac{2\sin ^2 \big(|w_i - w_j|\big)}{ r^2}\bigg) \:,  \: 
\end{align*}
with parameters $\theta = \{\sigma, r, \Sigma \}$. $\Sigma = DD^T$ has a low-rank form, and $D \in \R^{P \times m}$ is a (learned) matrix that captures information common to all images of each digit such as written style. 
This approach has multiple drawbacks. Firstly, given a set of $N$ images,
the above kernel gives rise to a dense matrix $\textbf{K}_{NN}$, where all latent variables are correlated with
each other. This necessitates either $\bigO(N^3)$ cost or non-trivial approximations, such as sparse GPs \citep{jazbec2020scalable} or assuming equally spaced $x_i$ enabling specialized matrix decompositions \citep{Casale2018GaussianAutoencoders}.
Statistically, the prior does not factorize across any subsets of the data.
Secondly, if new digits are added to the dataset, the $\Sigma$ matrix needs
be augmented with a new row and column, the new hyperparamters of $D$ must be
learned from scratch, and it is not ``amortized'' over digits.

In this work, we propose two simple (yet still unexplored) changes that alleviate the aforementioned issues.
%
%
We start by partitioning the dataset into digit specific subsets  $\{\Xb, \Yb, \Zb \} = \cup_{p=1}^P \{\Xb_p, \Yb_p, \Zb_p \}$ and denote partitions by
\begin{align*}
    \Xb_p &= \{\xb_i|d_i=p\} \in \R^{Q \times D}, \\
    \Yb_p &= \{\yb_i|d_i=p\} \in \R^{Q \times K}, \\
    \Zb_p &= \{\zb_i|d_i=p\}  \in \R^{Q \times L}.
\end{align*}
Instead of assuming a single correlated prior over all $N$ latent variables, 
we assume $P$ separate  correlated priors over $Q$ latent variables each, that is, one prior for each
partition of the data. Secondly, we assume that each latent variable $\zb_i\in\R^L$ 
is composed of two parts. Temporarily dropping $i$ for clarity, we propose to have
\begin{align*}
    \zb =  \big( \, \underbrace{z^1,..., z^J}_{local}, \,\, \underbrace{z^{J+1},..., z^L}_{global}\,\big) \; ,
\end{align*}
where the \emph{local} variable $\zb_i^{1:J}$ is unique to the given image and the
\emph{global} variable $\zb_i^{J+1:L}$ is shared by all elements in the subset $\Zb_p$.
The global variable captures style information or other common features of
digit $d_i$ that are angle-agnostic. For example, for frames from videos of moving objects,
local variables could capture object position and location while global variables could capture
object color and shape.

We must construct a kernel for each latent dimension $l \in \{1,..., L\}$ satisfying the above criteria.
For the local latent channels $l\in \{1,...,J\}$, we specify the following kernel\footnote{$\delta_{d_i, d_j} = 1$ if $d_i = d_j$ and $0$ else.}
\begin{align*}
    k_{\theta}^{1} (\xb_i, \xb_j) = \delta_{d_i, d_j} \cdot \sigma^2 \exp\bigg(-\frac{2\sin ^2 \big(|w_i - w_j|\big)}{ r^2}\bigg)  \: .
\end{align*}
The first Kronecker delta term ensures that two images corresponding to different
digits have zero assumed latent variable similarity, while for two images of the same digit the
assumed similarity depends upon the difference in the rotation angles. This enforces
that the GP priors for each subset of digits are independent.

For the global latent channels $l\in \{J+1,...,L\}$, we wish to capture rotation-agnostic characteristics
of each image, that is, its style. To this end, a simple binary kernel is used
\begin{align*}
    k^{2}_{\theta} (\xb_i, \xb_j) = \delta_{d_i, d_j}.
\end{align*}
Thus, among rotated images of the same digit, the global latents have 
perfect correlation; for each global channel there is a single univariate 
distribution shared by all images of a single digit. Simultaneously, the global
latent variable of images in another subset is treated as independent
(this may also be viewed as the local kernel with
length scale $r\to\infty$).
Due to the kernel structure, there is now a separate generative model
for every digit instance $p_{\psi, \theta} (\Yb, \Zb | \Xb) = \prod_{p=1}^P p_{\psi} (\Yb_p | \Zb_p) p_{\theta}(\Zb_p | \Xb_p)$. Within each of the digit-specific generative models, working
with the GP prior is much less prohibitive as $Q \ll N$. 
Secondly, digit style is captured in the global latent variables
$\zb_i^{j+1:L}$ which can be estimated from images via amortization.
Such global information is no longer encoded in the generative model
hyperparameters $\Sigma$.
%
%
%
\\ \\
\textbf{Approximate Posterior}: 
exploiting the factorized structure of the generative model, 
we may consider the posterior of each subset $\Zb_p$ independently. 
Since the true posterior for latent variables $p_{\psi, \theta}(\Zb_p | \Yb_p, \Xb_p)$
is intractable, approximate inference is required. In VAEs, an inference network (with
parameters $\phi$) takes $\yb_q\in\Yb_p$ as input to  predict the mean and variance of a
mean-field approximate posterior of each latent encoding
which we denote as
\begin{align*}
\tilde{q}_{\phi} (\zb_{q} | \yb_{q}) = \prod_{l=1}^{L} \mathcal{N}\big(z_{q}^{l} | \mu_{\phi}^{l}(\yb_{q}), \sigma_{\phi}^{l}(\yb_{q})^2\big) \;, 
\end{align*}
and one possible approximate posterior is to use the product of the above factors over all $N$ latent variables.
Instead, closely following \cite{Pearce2019ThePixels}, we use $\tilde{q}_\phi(\cdot)$ to
replace only the intractable likelihood $p_{\psi}(\yb_{i} | \zb_{i})$ in the
exact posterior. This gives rise to the following approximate posterior
\begin{align*}
    q(\Zb_p | \Yb_p, \Xb_p, \phi, \theta) := \frac{\prod_{q=1}^Q \tilde{q}_{\phi} (\zb_{q} | \yb_{q}) \cdot p_{\theta}(\Zb_p | \Xb_p)}{Z_{\phi, \theta} (\Yb_p, \Xb_p)}.
\end{align*}
The conjugacy of the Gaussian prior and (approximate) Gaussian likelihoods yields a closed-form solution for the normalizing constant $Z_{\phi, \theta}(\Yb_p, \Xb_p)$.  Moreover, the approximate posterior $q(\Zb_p | \cdot)$ is mathematically equivalent to a product of $J$ exact GP posteriors of $Q$ points (one GP for each angle latent channel) and $L - J$ univariate Gaussian distributions that are common to all $Q$ elements in the subset. Exact derivations are given in Appendix \ref{sec:derivations}.

Finally, the \ourmodel ELBO has the form: 
\begin{align*}
    \log p(\Yb | \Xb)  \ge  \sum_{p=1}^P \E_q \bigg[\sum_{q=1}^Q \log p_{\psi}(\yb_{q} | \zb_{q}) - \log \tilde{q}_{\phi} (\zb_{q} | \yb_{q}) \bigg] + \log Z_{\phi, \theta}(\Yb_p, \Xb_p) .
\end{align*}
Due to its factorization across digit subsets, and the assumption that $Q \ll N$, exact GP inference (with approximate likelihoods) is feasible, resulting in $\bigO(PQ^3)$ complexity for one epoch. Additionally, training can be done in mini-batches of digit subsets $\{\Yb_p, \Xb_p \}$, hence the ELBO does not require the whole dataset in memory. 
In cases where $Q$ is large or $\Yb_p$ do not fit into memory, the factorized kernel we propose
may be combined with a sparse GP-VAE method \citep{ashman2020sparse, jazbec2020scalable} to further reduce
computational complexity.

\section{Experiments}
\label{sec:experiments}


We follow the experimental setup from \citet{Casale2018GaussianAutoencoders} in conditionally generating rotated images of MNIST handwritten digits \citep{lecun1998gradient}.
The dataset consists of $P = 400$ different instances of the digit 3 at $Q = 16$ different angles each, resulting in a total of $N = 6400$ possible combinations.
From these combinations, $N_{train} = 4050$ images are used for training and $N_{test} = 270$ for testing.
We choose the same network architecture as the one in \citet{Casale2018GaussianAutoencoders} for all models (see Appendix~\ref{sec:details} for implementation details).
For running the baselines, we used the code from \citet{jazbec2020scalable}. Moreover, we make use of the GECO algorithm \citep{Rezende2018TamingVAEs} to train our \ourmodel model, as it improves training stability. Our code is made available at \url{https://github.com/metodj/FGP-VAE}.

\begin{table}
\setlength{\tabcolsep}{10pt}
\centering
\caption{Results on the rotated MNIST digit 3 dataset. Reported here are mean values together with standard deviations based on 5 runs. We see that our proposed model outperforms the baselines while still being more scalable than the \citet{Casale2018GaussianAutoencoders} model. $P$ represents the number of unique digits, $Q$ the number of rotations for each digit and $m$ the dimension of the low-rank matrix in the GP kernel used in \cite{Casale2018GaussianAutoencoders}.}
\resizebox{\linewidth}{!}{
\begin{tabular}{l l l l}
\toprule
 & \textbf{MSE} & \textbf{GP complexity}   & \textbf{Time/epoch [s]}\\
\midrule
\textbf{CVAE} \citep{sohn2015learning} & $0.0796 \pm 0.0023$  & - & $0.39 \pm 0.01$ \\[0.08cm]
  \textbf{GP-VAE} \citep{Casale2018GaussianAutoencoders} & $0.0370 \pm 0.0012$ & $\bigO(PQ^3m^2)$ & $19.10 \pm 0.66$\\[0.08cm]
\textbf{\ourmodel} (ours) & $0.0284 \pm 0.0004$ & $\bigO(PQ^3)$ & $1.41 \pm 0.08$\\[0.08cm]
\bottomrule
\end{tabular}
}
\label{table:rot_MNIST_main}
\end{table}



\begin{figure}
    \centering
    \subfigure[Conditional generations]{
    \label{fig:mnist_images}
    \includegraphics[width=0.4\linewidth]{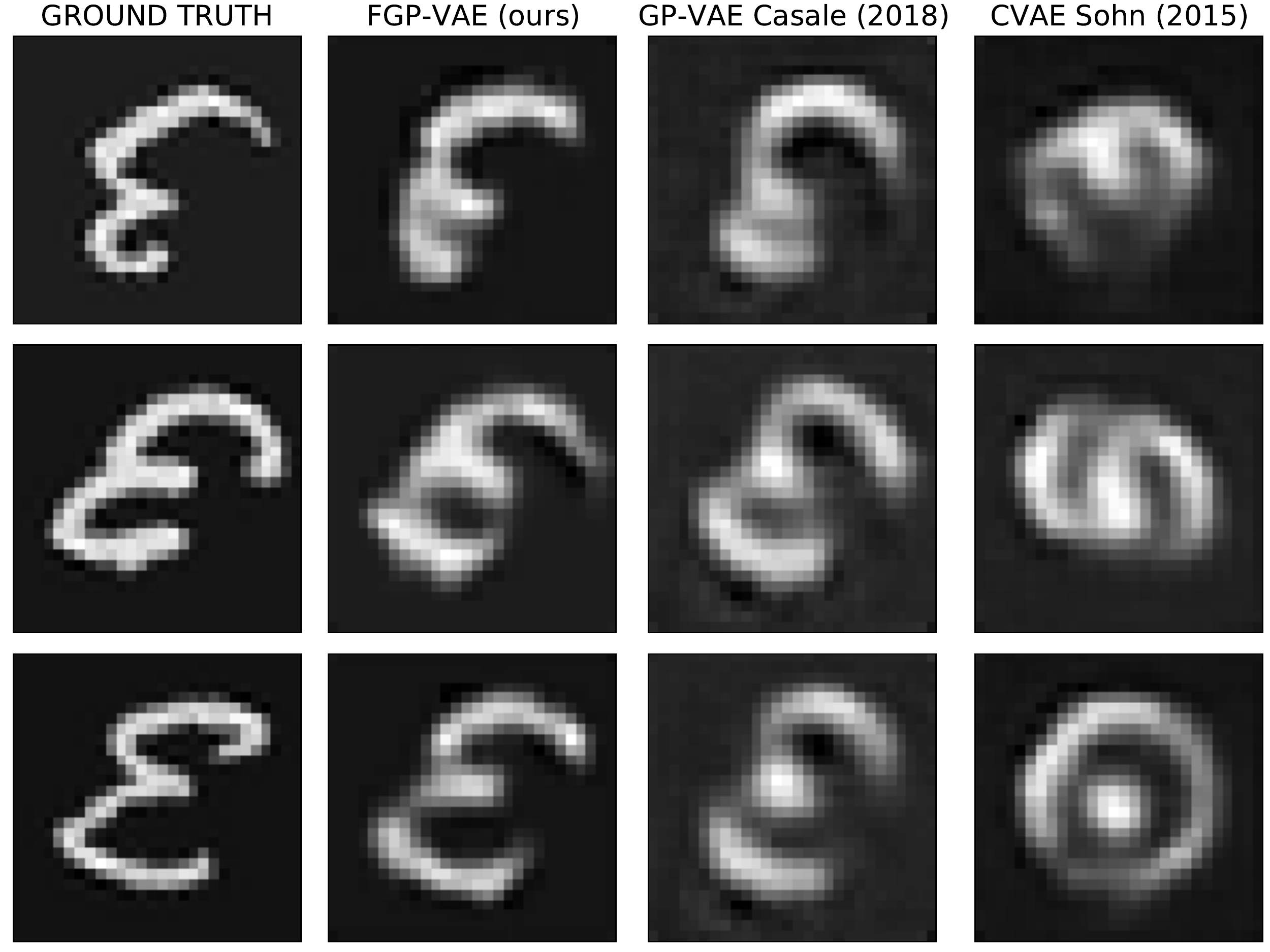}}%
    \hfill
    \subfigure[Scaling properties]{
    \label{fig:mnist_scaling}
    \includegraphics[width=0.57\linewidth]{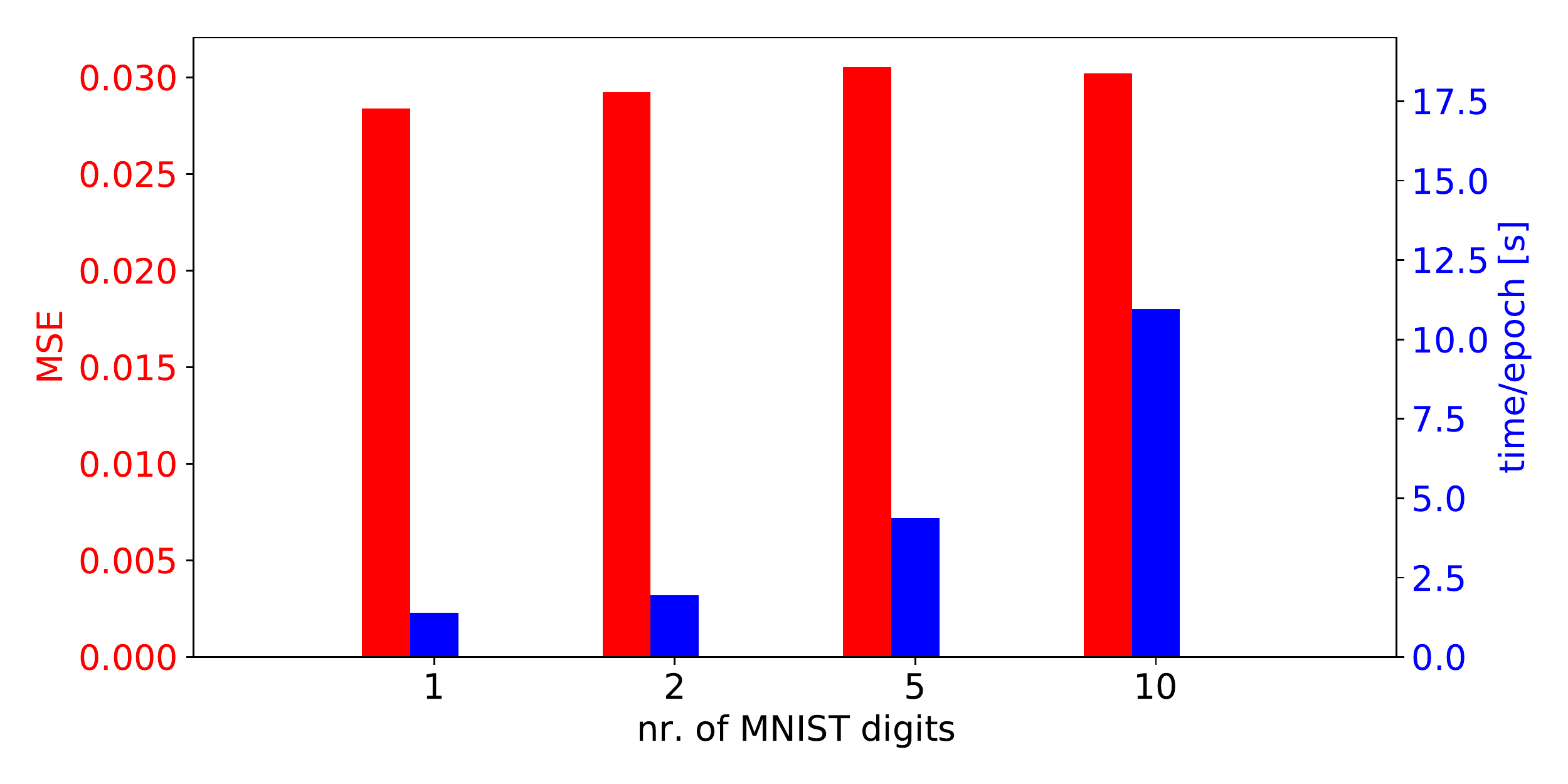}}%
    \caption{(a) Conditionally generated rotated MNIST images. The generations of our proposed model are qualitatively more faithful to the ground truth. (b) Performance and runtime of our proposed model on differently sized subsets of the MNIST dataset, including the full set. We see that the performance stays roughly the same, regardless of dataset size, while the runtime grows linearly as expected. The size of each dataset equals $4050 \times \textrm{nr. of MNIST digits}$.}
    \label{fig:combined_images_scaling}
\end{figure}

\paragraph{Reconstruction performance.}
We see qualitatively in Figure~\ref{fig:mnist_images} and quanitatively in Table~\ref{table:rot_MNIST_main} that our proposed \ourmodel clearly outperforms the non-correlated CVAE model \citep{sohn2015learning} and performs comparably to the non-factorized GP-VAE \citep{Casale2018GaussianAutoencoders}.
However, our proposed model is an order of magnitude faster than the non-factorized GP-VAE and reaches runtimes that are almost as fast as the CVAE.

\paragraph{Scaling behavior.}
We additionally studied the scaling of our proposed model to differently sized subsets, including the full dataset.
(We do not compare against the other GP-VAE model here, since we did not manage to scale it to this larger dataset).
We can see in Figure~\ref{fig:mnist_scaling} that our proposed model does not deteriorate in performance when scaling the dataset, while the runtime scales as gracefully as theoretically predicted (see Sec.~\ref{sec:method}).

\paragraph{Extrapolation in the digit space.} An additional attractive feature of the factorized model
is that it can (unlike past works) extrapolate beyond digits observed in the training data.
Past works only considered images seen during training and generated them at new angles.
The \ourmodel can also generate arbitrary rotations for previously unseen digits given either a single example (sample the posterior of global latents), or even generated randomly (sample the prior of global latents).
This is a consequence of our disentangled (and arguably much simpler) GP kernel. 
The \ourmodel achieved an MSE of $0.0316 \pm 0.0005$ for the extrapolation
experiment, which is only slightly worse than the $0.0284 \pm 0.0004$ for
the non-extrapolation version of the experiment. This demonstrates the
\ourmodel's strong extrapolation ability in the digit space. Generated
images from the extrapolation experiment are shown in Appendix~\ref{sec:extrapolate-figure}.
We further elaborate on the extrapolation properties of GP-VAE models in Appendix~\ref{sec:amort-aux-data}.

\section{Conclusion}
\label{sec:conclusion}

We have proposed a novel method for improving the scalability of GP-VAE models in settings where the auxiliary data consists of several independent features.
Our method factorizes the latent GP kernel across the different data features in different latent dimensions, leading to a large reduction in inference time complexity.
We have shown that our model is faster than existing non-factorized approaches in practice, while yielding a comparable predictive performance and offering more general extrapolation properties.
In future work, it would be interesting to study the combination of our method with the recently proposed sparse GP-VAE approaches \citep{jazbec2020scalable, ashman2020sparse}, to reduce the inference time even further.


\bibliography{refs}

\appendix

\counterwithin{figure}{section}
\counterwithin{table}{section}

\section{Implementation details}
\label{sec:details}

For the rotated MNIST experiment described, we used the same neural networks architectures as in \cite{Casale2018GaussianAutoencoders}: three convolutional layers followed by a fully connected layer in the inference network  and vice-versa in the generative network. For more details, see Table \ref{tab:mnist_parameters}.

\begin{table}[h!]
    \centering
    \caption{Neural networks architectures for the MNIST experiment.}
    \begin{tabular}{lc}
    \toprule
        Parameter & Value \\
        \midrule
         Nr. of CNN layers in inference network &  3 \\
         Nr. of CNN layers in generative network &  3 \\
         Nr. of filters per CNN layer & 8 \\
         Filter size & $3 \times 3$ \\
         Nr. of feedforward layers in inference network & 1 \\
         Nr. of feedforward layers in generative network & 1 \\
         Activation function in CNN layers & ELU \\
         Dimensionality of latent space (L) & 16 \\
         Number of latent channels for angle info (J) & 8 \\
         \bottomrule
    \end{tabular}
    \label{tab:mnist_parameters}
\end{table}
The \ourmodel model is trained for 1000 epochs with a batch size of 220 images (20 digits subsets, each with 11 rotations). The Adam optimizer \citep{Kingma2014Adam:Optimization} is used with its default parameters and a learning rate of 0.001. Moreover, the GECO algorithm \citep{Rezende2018TamingVAEs} is used for training our \ourmodel model in this experiment. The reconstruction parameter in GECO was set to $\kappa = 0.020$ in all reported experiments. 

GP parameters are kept fixed throughout training for \ourmodel. The amplitude is set to $\sigma = 1$ and the length scale to $r = 1$. For the baseline GP-VAE model \citep{Casale2018GaussianAutoencoders}, GP parameters are optimized during training as proposed in \cite{Casale2018GaussianAutoencoders}.

\section{Derivations}
\label{sec:derivations}
To simplify the notation, we use here one latent channel per feature set, i.e. $J = 1$ and $L = 2$. For $\zb_q \in \Zb_p$, we thus have that $\zb_q = [z_q \; z_p]^T$, where we drop latent channel superscripts for clarity. With $z_q$ we denote a local latent variable that is specific to the $q$-th rotation, while $z_p$ represents a global latent variable that is shared among all rotations of the $p$-th digit. Further, let $\zb_p^1 = [z_1 \dots z_Q]^T \in \R^{Q}$ contain all local latent variables in $\Zb_p$, and $\Kb_p = k_{\theta}^1(\Xb_p, \Xb_p)$.  For notational convenience, let $\mu_{q,l} := \mu_{\phi}^{l}(\yb_{q})$ and $\sigma_{q,l} :=\sigma_{\phi}^{l}(\yb_{q})$. We proceed as

\begin{multline*}
    Z_{\phi, \theta} (\Yb_p, \Xb_p) = \int \prod_{q=1}^Q \tilde{q}_{\phi} (\zb_{q} | \yb_{q}) \cdot  p_{\theta}(\Zb_p | \Xb_p) \: d\Zb_p =  \\ \int \prod_{q=1}^Q \mathcal{N}(z_{q} | \mu_{q,1}, \: \sigma_{q ,1}^2) \: \mathcal{N}(z_p | \mu_{q,2}, \: \sigma_{q ,2}^2) \cdot \mathcal{N}(\zb_p^{1} | \mathbf{0}, \Kb_p) \:  \mathcal{N}(z_p| 0, 1) \: d\Zb_p = \\ \int \prod_{q=1}^Q \mathcal{N}(z_{q} | \mu_{q,1}, \: \sigma_{q ,1}^2) \: \mathcal{N}(\zb_p^{1} | \mathbf{0}, \Kb_p) \: d\zb_p^{1} \cdot \int \prod_{q=1}^Q \mathcal{N}(z_p | \mu_{q,2}, \: \sigma_{q ,2}^2) \:  \mathcal{N}(z_p | 0, 1) \: dz_p \: .
\end{multline*}
By exploiting the symmetry of Gaussian distribution, $\mathcal{N}(z| \mu, \sigma) = \mathcal{N}(\mu |z, \sigma)$, the first integral equals a marginal GP likelihood in the standard GP regression with inputs $\Xb_p$ and outputs $\boldsymbol{\mu}_p := [\mu_{1,1} \dots \mu_{Q,1}]^T \in \R^Q$ with (heteroscedastic) noise $\boldsymbol{\sigma}_p := [\sigma_{1,1} \dots \sigma_{Q,1}]^T \in \R^Q$. Combining the same symmetry property with a formula for conjugate posterior parameters for Gaussian likelihood with known heteroscedastic variance yields the following expression for the second integral
\begin{align*}
\int \prod_{q=1}^Q \mathcal{N}(z_p | \mu_{q,2}, \: \sigma_{q ,2}^2) \:  \mathcal{N}(z_p | 0, 1) \: dz_p = \frac{\mathcal{N}(0 | 0, 1) \prod_{q=1}^Q \mathcal{N}(0 | \mu_{q,2}, \: \sigma_{q ,2}^2)}{\mathcal{N}(0 | \bar{\mu}_{2}, \bar{\sigma}_{2}^2)} \: ,
\end{align*}
where
\begin{align*}
\bar{\sigma}_2^2 = \big(1 + \sum_{q=1}^Q \frac{1}{\sigma_{q, 2}^2}\big)^{-1}\: , \hspace{10pt}
    \bar{\mu}_2 = \bar{\sigma}_2^2 \: \sum_{q=1}^Q \frac{\mu_{q, 2}}{\sigma_{q, 2}^2} \; .
\end{align*}
Similarly, a closed form for the approximate posterior can be obtained as
\begin{multline*}
     q(\Zb_p | \Yb_p, \Xb_p, \phi, \theta) = \frac{\prod_{q=1}^Q \tilde{q}_{\phi} (\zb_{q} | \yb_{q}) \cdot p_{\theta}(\Zb_p | \Xb_p)}{Z_{\phi, \theta} (\Yb_p, \Xb_p)}  = \\[2ex]
     \underbrace{\frac{\prod_{q=1}^Q \mathcal{N}(z_{q} | \mu_{q,1}, \: \sigma_{q ,1}^2) \: \mathcal{N}(\zb_p^{1} | \mathbf{0}, \Kb_p)}{\int \prod_{q=1}^Q \mathcal{N}(z_{q} | \mu_{q,1}, \: \sigma_{q ,1}^2) \: \mathcal{N}(\zb_p^{1} | \mathbf{0}, \Kb_p) \: d\zb_p^{1}}}_\text{(exact) GP posterior for $\{\Xb_p, \boldsymbol{\mu}_p, \boldsymbol{\sigma}_p \}$} \cdot \underbrace{\frac{\prod_{q=1}^Q \mathcal{N}(z_p | \mu_{q,2}, \: \sigma_{q ,2}^2) \:  \mathcal{N}(z_p | 0, 1) }{\int \prod_{q=1}^Q \mathcal{N}(z_p | \mu_{q,2}, \: \sigma_{q ,2}^2) \:  \mathcal{N}(z_p | 0, 1) \: dz_p{}}}_\text{$= \mathcal{N}(z_p | \bar{\mu}_2 , \: \bar{\sigma}_2^2 )$, Gaussian posterior} \: .
\end{multline*}
Finally, an \ourmodel ELBO can be derived as follows using the standard steps:
\begin{multline*}
    \log p(\Yb | \Xb)  \ge \sum_{p=1}^P \int \log \frac{p_{\psi, \theta}(\Yb_p, \Zb_p | \Xb_p)}{q(\Zb_p | \cdot)} q(\Zb_p | \cdot)d\Zb_p = \\
    \sum_{p=1}^P \int \log \bigg( \frac{ p_{\psi} (\Yb_p | \Zb_p)\cdot p_{\theta}(\Zb_p | \Xb_p) \cdot Z_{\phi, \theta}(\Yb_p, \Xb_p)}{\prod_{q=1}^Q \tilde{q}_{\phi} (\zb_{q} | \yb_{q}) \cdot p_{\theta}(\Zb_p | \Xb_p)} \bigg) q(\Zb_p | \cdot)d\Zb_p = \\ \sum_{p=1}^P \E_q \bigg[\sum_{q=1}^Q \log p_{\psi}(\yb_{q} | \zb_{q}) - \log \tilde{q}_{\phi} (\zb_{q} | \yb_{q}) \bigg] + \log Z_{\phi, \theta}(\Yb_p, \Xb_p) .
\end{multline*}

\section{Amortization of auxiliary data in GP-VAE models}
\label{sec:amort-aux-data}
Auxiliary data $\Xb$ is crucial in applications of GP-VAE models as it represents the data over which a GP prior is placed.\footnote{Auxiliary data in a GP-VAE corresponds to independent variables in a GP regression.} While it is often fully observed, there are cases where auxiliary data is not given (or is only partially observed). In such instances, the authors in \cite{Casale2018GaussianAutoencoders} rely on the GP-LVM \citep{Lawrence2004GaussianData} to learn the missing parts of the auxiliary information. Such an approach solves the issue of (partly) unobserved $\Xb$ in an elegant way, however by doing so, the extrapolation ability of GP-VAE models is diminished. Suppose we want to generate new views or angles for previously unseen digits or objects. In that case, we need to re-run the training optimization so that the respective GP-LVM vectors are obtained. Note that GP-LVM vectors correspond to rows in the low-rank matrix $\Sigma \in \R^{P \times m}$ that is part of the GP kernel proposed in \cite{Casale2018GaussianAutoencoders}.

Another way of endowing GP-VAE models with the extrapolation ability, besides considering factorized (and simpler) GP priors as done in our \ourmodel, would be to amortize the GP-LVM information using a \emph{representation network} $r_{\zeta}: \R^K \to \R^m$, similar to what is done in \cite{AliEslami2018NeuralRendering}.  The representation for the $p$-th digit instance is then
\begin{align*}
    \mathbf{d}_p = f\big(r_{\zeta}(\yb_{1}), \: \dots, \: r_{\zeta}(\yb_{Q})\big) \in \R^m \: , 
\end{align*}
where $\Yb_p = [\yb_1 \dots \yb_Q]^T$, and $f$ is a chosen aggregation function, for instance, a sum or a mean. Instead of GP-LVM vectors, the parameters of the representation network $\zeta$ would be learned jointly with the rest of GP-VAE parameters.


\section{Extrapolation in the digit space}\label{sec:extrapolate-figure}
\begin{figure}[!ht]
\centering
\includegraphics[scale=0.40]{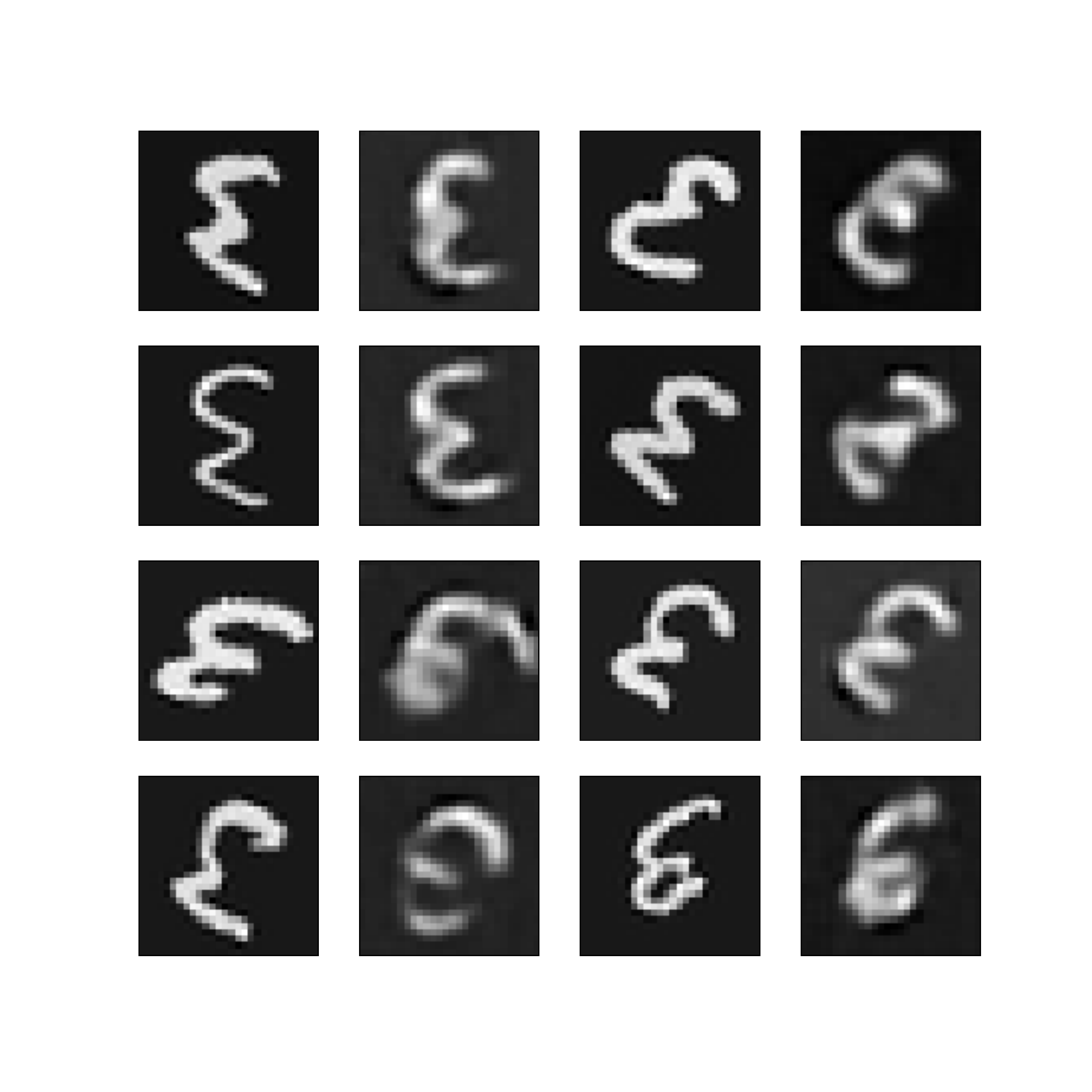}
\caption{Ground truths (columns 1 and 3) and generated images (columns 2 and 4) using \ourmodel for new digit instances (not seen during training in any angle). To generate the rotations for each new digit in the test phase, 11 context images were given to the model.}
\label{fig:rot_MNIST_reGPVAE_extrapolate}
\end{figure}

\end{document}